**Title:** Extracting effective solutions hidden in large language models via generated comprehensive specialists: case studies in developing electronic devices

**Author:**

Hikari Tomita[1,#], Nobuhiro Nakamura[2,3#], Shoichi Ishida[4], Toshio Kamiya[2], Kei Terayama[1,2,4*]

**Affiliation:**

[1]School of Science, Yokohama City University, Tsurumi-ku, Yokohama 230-0045, Japan

[2]MDX Research Center for Element Strategy, Institute of Science Tokyo, Yokohama, 226-8501, Kanagawa, Japan.

[3]Materials Integration Laboratories, AGC Inc, Yokohama, 230-0045, Japan

[4]Graduate School of Medical Life Science, Yokohama City University, Tsurumi-ku, Yokohama 230-0045, Japan

#equally contributed

*Corresponding author(s). E-mail(s): terayama@yokohama-cu.ac.jp

**Abstract:**

Recently, many studies have increasingly explored the use of large language models (LLMs) to generate research ideas and scientific hypotheses. However, real-world research and development often require solving complex, interdisciplinary challenges where solutions may not be readily found through existing knowledge related to the problem. Therefore, it is desirable to leverage the vast, comprehensive knowledge of LLMs to generate effective, breakthrough solutions by integrating various perspectives from other disciplines. Here, we propose SELLM (Solution Enumeration via comprehensive List and LLM), a framework leveraging LLMs and structured guidance using MECE (Mutually Exclusive, Collectively Exhaustive) principles, such as International Patent Classification (IPC) and the periodic table of elements. SELLM systematically constructs comprehensive expert agents from the list to generate cross-disciplinary and effective solutions. To evaluate SELLM's practicality, we applied it to two challenges: improving light extraction in organic light-emitting diode (OLED) lighting and developing electrodes for next-generation memory materials. The results demonstrate that SELLM significantly facilitates the generation of effective solutions compared to cases without specific customization or effort, showcasing the potential of SELLM to enable LLMs to generate effective solutions even for challenging problems.

# 1. Introduction:

In recent years, various methods leveraging large language models (LLMs)[1–3] for generating research ideas[4–7], solutions[8], and hypotheses[9–11] to solve specific problems—traditionally performed by humans—have been actively studied and developed. To enable problem-solving based on specialized knowledge, approaches including fine-tuning with domain-specific datasets[12,13] and utilizing retrieval-augmented generation (RAG)[8,14,15] have been explored to enhance expertise. Furthermore, research has also explored enhancing the diversity and feasibility of generated solutions by facilitating interactions between multiple expert models with domain-specific knowledge[6,16,17]. The applications of LLM-based approaches have also been reported in the fields of materials science and materials development[18,19]. Additionally, there are reports of using knowledge graphs to harness networks of specialized knowledge for generating research ideas and hypotheses[5,20]. More recently, studies have even demonstrated the ability to draft entire research papers using such techniques[21].

However, the development of advanced methodologies for generating effective solutions to intrinsically challenging problems is still in its early stages. By "intrinsically challenging problems," we refer to issues that require the skillful integration of knowledge and methodology from seemingly unrelated fields to address effective solutions. Prominent examples of such problems and their solutions include CRISPR-Cas9[22] for genetic engineering[23] (Nobel Prize 2020), Green Fluorescent Protein (GFP)[24] for live-cell imaging[25] (Nobel Prize 2008), and functional Magnetic Resonance Imaging (fMRI) for brain activity mapping[26–28]. The application of Indium Gallium Zinc Oxide (IGZO) technology from flat panel displays to Dynamic Random Access Memory (DRAM)[29,30], which is treated as a target problem in this study, also serves as a representative example of such challenges. Addressing these types of problems is not straightforward with existing methodologies, because identifying the appropriate domain of expertise distant from the target problem is inherently difficult. Some approaches have been proposed to utilize knowledge from fields distant from the target problem[15,20]; however, the challenge of missing critical knowledge or fields essential for problem-solving has not been sufficiently addressed.

In this study, we propose a brute-force approach called SELLM (Solution Enumeration via comprehensive List and LLM), designed to generate valuable solutions to inherently challenging problems using LLMs. The core principle of SELLM is to construct a comprehensive set of domain-specific "experts" using LLMs, with each expert proposing specialized solutions from the perspective of its respective field. If a comprehensive set of experts with appropriate granularity is established, it becomes possible to generate effective solutions that leverage seemingly unrelated knowledge and methodologies, without omissions. To achieve them, SELLM utilizes structured lists of MECE (Mutually Exclusive, Collectively Exhaustive) principles, such as the International Patent Classification (IPC) system or the periodic table of elements, to generate solutions that incorporate domain-specific expertise.

To evaluate the effectiveness of the proposed method, we applied SELLM to two challenging problems: improving light extraction in organic light emitting diode (OLED) lighting[31,32] and developing electrodes for next-generation memory materials[30]. It should be noted that while the solutions to these problems may seem straightforward once presented, coming up with them requires significant expertise of different fields, making these inherently challenging problems. We demonstrated that SELLM, by utilizing IPC subclass lists and chemical element lists, facilitates the generation of effective solutions compared to cases where no specific customization or effort is applied. These results highlight the potential of effectively leveraging LLMs to propose essential solutions by integrating knowledge from seemingly unrelated fields.

## 2. Result
### 2.1. Overview of the proposed method and evaluation strategy

Figure 1 illustrates the framework of SELLM. The input consists of a description of the problem to be solved and a list of knowledge elements with explanations for each element. The list is expected to maintain an appropriate level of granularity suitable for its field of expertise while ensuring sufficiently broad coverage. Additionally, minimizing overlap between elements helps to avoid unnecessary redundancy. Subsequently, an expert is generated for each element using the technique of role-play prompting[33] from the provided list. These experts generate solutions to the given problem based on their respective expertise. Further details are provided in the Methods section.

To assess the appropriateness of the generated solutions, three evaluation methods are employed: similarity-based evaluation (SBE), keyword-based evaluation (KBE), and human-based evaluation (HBE). In SBE, leveraging the LLM-as-a-Judge[34] approach, the similarity between the generated solutions and the "reference solution", which is described in the literature, is rated on a scale from 1 to 10. For KBE, key terms essential to problem-solving are scored by counting the occurrence of these terms. In HBE, some of the solutions with diverse SBE scores are further evaluated by human experts to determine whether they constitute genuinely effective solutions, irrespective of alignment with the reference solution. The details are described in the Methods section.

Here, we introduce two challenging problems addressed in this study and their respective solutions briefly. The first problem is the efficient extraction of light from OLED devices used for lighting. In the 2010s, a major issue was that the refractive indices of the organic layers and transparent electrodes constituting the organic EL devices were high (ranging from 1.7 to 2.0), resulting in only about 20% of the emitted light being extracted externally[35–37] (Figure 2(a)). Despite significant efforts by various industries and academia to resolve this issue, developing a low-cost and stable manufacturing method proved challenging[38–40]. Ultimately, an excellent solution was proposed by combining glass film formation techniques involving lens glass materials used in optical lenses and glass frit paste employed for forming front dielectric layers in plasma televisions (Figure 2(b))[31]. This technology is known as KIWI Technology. Since the patents and publications regarding this solution have been publicly available, recent pre-trained models would have been trained on this information. However, as demonstrated later, generating the essential solutions without specific guidance remains challenging.

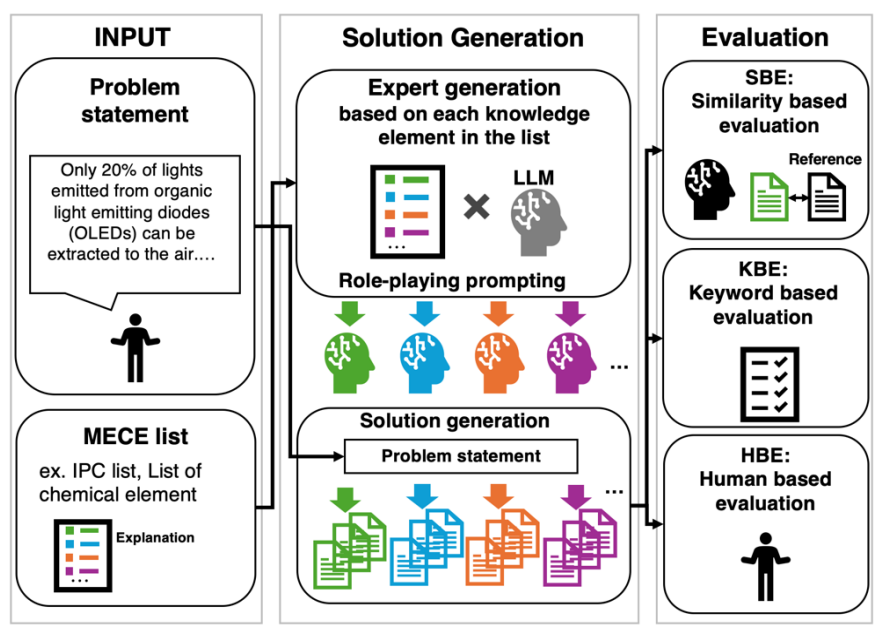

Figure 1. Overview of the proposed method and its evaluation.

The second problem involves addressing the contact resistance issue in IGZO-based thin-film transistors (IGZO-TFTs), which are promising candidates for next-generation memory materials capable of achieving high speed and large capacity. These devices use IGZO, an amorphous oxide semiconductor, as the primary material[30] (Figure 2(c)). The practical application of IGZO in memory devices requires the implementation of fine wiring and electrodes, but challenges such as mobility reduction and increased power consumption due to contact resistance have emerged. Various approaches have been proposed to tackle these issues[41–43]. Recently, a solution was found using palladium (Pd), a catalytic metal with high hydrogen permeability and the ability to dissociate hydrogen molecules[30]. This solution, detailed in a paper published in 2024, is unlikely to have been included as training data for the pre-trained models used in this study.

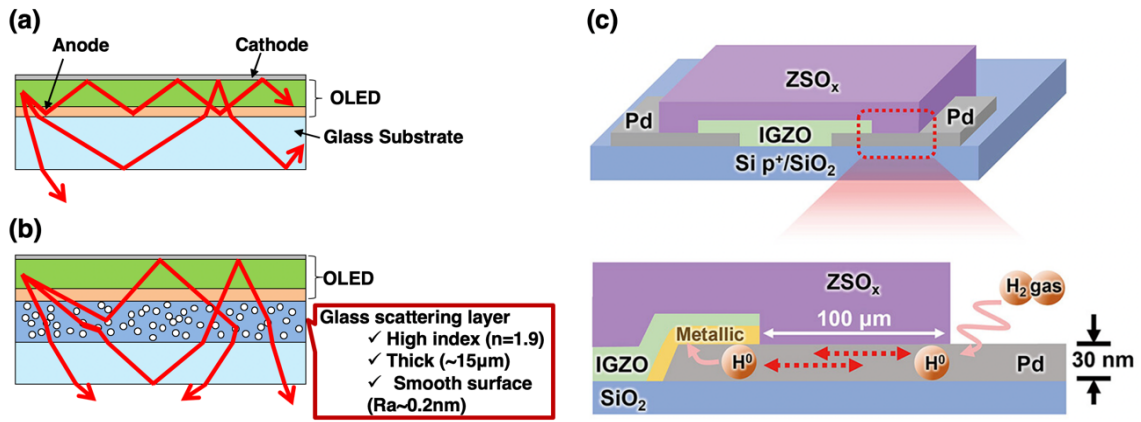

Figure 2. An overview of the challenging problems and solutions addressed in this study. (a) Efficient light extraction from OLED devices for lighting was hindered by the high refractive indices of organic layers and transparent electrodes. (b) KIWI technology resolved this issue by combining glass film formation techniques involving lens glass materials used in optical lenses and glass frit paste employed for forming front dielectric layers in plasma televisions. (c) An overview of the contact resistance issue in IGZO-TFTs and its solution using palladium electrodes. Adapted with permission from Shi et al.[30] under the terms of a Creative Commons Attribution License 4.0 (CC BY).

## 2.2. Generating solutions to the light extraction problem in OLED lighting

We conducted a validation to evaluate SELLM's capability to generate solutions for the light extraction problem. The problem statement, reference solution, and keywords are provided in Table 1, while the list of subclasses from the IPC sections B, C, F, G, and H was used. For comparison, solutions were generated using the same problem statement without any specific adjustments (Standard). Considering the balance between performance and cost, OpenAI's gpt-4o-2024-08-06 (GPT-4o) as an representative LLM was utilized for the generation of experts and solutions. Details of the explanatory descriptions and generation conditions are described in the Methods section.

Table 1. Problem statement, reference solution, and keywords for the light extraction problem.

| Problem Statement | Only 20% of lights emitted from organic light emitting diodes (OLEDs) can be extracted to the air. It is because organic layers and the transparent electrode of OLEDs have high refractive indices (~1.9) and thus the emitted lights are confined by total reflection at two interfaces: 1. the interface of the transparent electrode and the glass substrate and 2. the interface of the glass substrate and the air. To extract a large amount of light from the device, applying a scattering layer with high refractive index on the glass, and fabricating an OLED on the scattering layer would be happy. In this case, the refractive index of the scattering layer matrix needs to be equal or higher than 1.9. Furthermore, the surface of the scattering layer must be smooth enough because short circuits easily occur between electrodes, if the surface of the scattering layer is rough. Please explain how to make the scattering layer, satisfying the above requirements. |
|---|---|

| Reference solution[31] | One solution is fabricating the glass scattering layer using glass frit and ceramic particles. First, frit paste is prepared by mixing highly refractive index glass frit, ceramic particles with a different refractive index, and a resin and a solvent, such as ethyl cellulose, terpineol respectively. The frit paste is printed on a glass substrate by screen printing. Then the substrate is heated to decompose the binder, and fired near the softening point. Finally a high refractive index scattering layer with a smooth surface can be formed on the glass substrate. |
|---|---|
| Keywords | Group1: frit paste, glass frit, enamel, glaze, and glass powder. Group2: firing, fired, sintering, sintered, and sinter. |

Figures 3a and 3b show the distribution of evaluation scores for SBE and KBE, respectively. The correct solution for SBE and the keywords for KBE are listed in Table X. Figure 3a demonstrates that, while the Standard approach yields solutions a maximum score of 6, SELLM successfully generates ones with high scores of 8 and 9, which are closer to the correct solution. This is particularly intriguing, given that KIWI technology is publicly available and likely part of the LLM training data, yet the Standard approach fails to generate viable solutions without specific adjustments. Similarly, Figure 3b shows that the Standard approach produces solutions with a maximum KBE score of only 1, whereas SELLM generates solutions with a KBE score of 2. Examples of high-scoring solutions for both SBE and KBE are provided in Table 2.

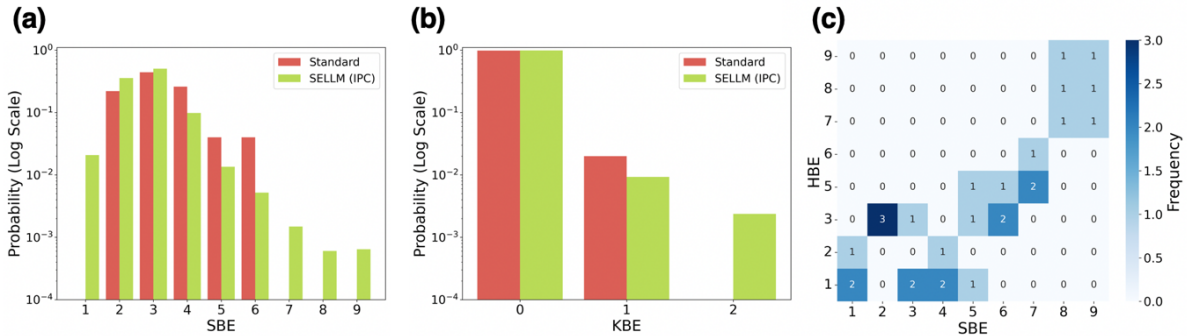

Figure 3. Evaluation of solutions generated by SELLM for the light extraction problem. (a) Distribution of SBE scores for solutions generated by SELLM and the Standard approach, presented on a logarithmic probability scale. (b) Distribution of KBE evaluation scores, also presented on a logarithmic probability scale. (c) Relationship between SBE and HBE scores for a subset of generated solutions evaluated using HBE.

Table 2: Examples of solution generated by SELLM with high SBE and KBE scores by SELLM for the light extraction problem.

| Sub-class | Generated solution | SBE | KBE |
|---|---|---|---|
| C03C | Formulate a scattering layer using a high-refractive-index glass frit. The frit, consisting of finely ground glass, is mixed with a binding agent to form a paste, which is then spread on a substrate and fired to form a dense, uniform layer. The glass composition should include | 9 | 2 |

| | high-index elements such as lead or bismuth oxides to enhance refractive properties. Post-firing, the layer is polished to achieve the required smoothness. This solution is in line with IPC code C03C 6/00, which focuses on glass batch compositions and frit technologies for modifying optical characteristics. | | |
|---|---|---|---|
| F26B | A method to create a scattering layer involves using a high-refractive-index glass frit that can be sintered onto the substrate. The glass frit is applied as a paste, then dried and sintered to form a dense, smooth layer. Conductive drying (IPC F26B 3/18) can be employed to ensure uniform temperature distribution during the drying phase, which is critical to prevent surface irregularities. The sintering process consolidates the glass frit into a uniform optical layer, enhancing light extraction while maintaining a smooth surface to prevent electrical shorts in the OLED. | 9 | 2 |

Next, evaluation using HBE was conducted. Among the solutions generated by SELLM, three solutions were selected for each SBE score from 1 to 10 and subsequently evaluated using HBE. Table S1 lists the evaluated solutions along with their respective HBE scores. Figure 3c illustrates the relationship between SBE and HBE scores. Overall, there is a correlation between SBE and HBE, suggesting that SBE, as assessed by the LLM, reasonably reflects the effectiveness of the solutions. As highlighted in Table 3, SELLM successfully generated highly accurate solutions with an HBE score of 9. For example, the second solution shown in Table 3 has an SBE score of 9 but an HBE score of 7, because, while it proposes a promising solution, it includes unnecessary steps in the process. Conversely, the third solution demonstrates an example where the HBE score is higher than the SBE score, achieving an HBE score of 9. These results indicate that the solutions generated by SELLM are generally deemed highly valid by human experts and that the evaluation strategy employed is working.

Table 3 Examples of solutions with relatively high HBE scores for the light extraction problem.

| Sub-class | Generated solution | SBE | KBE | HBE | Comment by human expert |
|---|---|---|---|---|---|
| C03B | A scattering layer can be manufactured by using a two-step thermal treatment process. First, a high refractive index glass frit, such as barium oxide-based glass, is prepared by melting the raw materials, as detailed in IPC code C03B 5/00, and then ground into fine particles. These particles are mixed with a binder to form a paste. The paste is applied to a glass substrate and initially heated at a lower temperature to evaporate the binder and allow the particles to adhere. A secondary higher temperature treatment follows, which melts and fuses the particles into a continuous layer. This process allows control over the layer's thickness and surface smoothness, preventing short circuits in OLEDs. | 9 | 1 | 9 | The idea of using a high-refractive-index glass frit paste to form a scattering layer during the firing process replicates the original inspiration. It also mentions glass materials for achieving a high refractive index. Although there is no mention of scattering particles, the bubbles present in the fired film act as substitutes for scattering particles, so there is no issue. |
| C03C | Formulate a scattering layer using a high-refractive-index glass frit. The frit, consisting of finely ground glass, is mixed with a binding agent to form a paste, which is then spread on a substrate and fired to form a dense, uniform layer. The glass composition should include high-index elements such as lead or bismuth oxides to enhance refractive properties. Post-firing, the layer is polished to achieve the required smoothness. This solution is in line with IPC code C03C 6/00, which focuses on glass batch compositions and frit technologies for modifying optical characteristics. | 9 | 2 | 7 | The idea of using a high-refractive-index glass frit paste to form a scattering layer during the firing process replicates the original inspiration but includes an unnecessary polishing process. There is a lack of understanding that a smooth surface can be achieved through the firing process. While there is no mention of scattering particles, the bubbles present in the fired layer act as substitutes for scattering particles, so there is no issue. |
| C23D | Design a vitreous scattering layer specifically for OLEDs by tailoring the enamel composition with barium or strontium oxides, which are known to increase the refractive index. The enamel is applied using a wet method (C23D 5/02) to maintain surface smoothness, essential for preventing electrical shorts. After application, a controlled firing process (C23D 9/00) ensures the enamel is clear and smooth, optimizing light extraction from the OLED. | 8 | 2 | 9 | The idea of using a high-refractive-index glass frit paste to form a scattering layer during the firing process replicates the original inspiration. It also mentions glass materials for achieving a high refractive index. Although there is no mention of scattering particles, the bubbles present in the fired film act as substitutes for scattering particles, so there is no issue. |

**2-3 Generating solutions to the problem of contact resistance in IGZO-TFT**

Figure 4 illustrates the solution generation results for the challenge in IGZO-TFT using SELLM. The problem statement is provided in Table 4, and the lists used include subclasses from the IPC classes B, C, F, G, and H, as well as a list of 83 elements ranging from hydrogen to bismuth. GPT-4o was also used for the generation of experts and solutions, considering the balance between performance and cost. Figures 4a and 4b show the distribution of evaluation scores for SBE and KBE, respectively. The correct solutions for SBE and the keywords for KBE are detailed in Table 4. From Figure 4a, it is evident that the maximum SBE score for the Standard approach is 4, whereas SELLM successfully generates solutions with significantly higher scores. Moreover, the element list tends to produce higher-scoring solutions compared to the IPC subclass list. Similarly, Figure 4b indicates that SELLM generates solutions with higher KBE scores, further suggesting that the element list is more effective than the IPC subclass list, as seen in the SBE results. Examples of generated solutions with high SBE and KBE scores are shown in Table 5. These results indicate that SELLM is also effective in addressing the problem of contact resistance in IGZO-TFTs, as it generates solutions close to the reference solution.

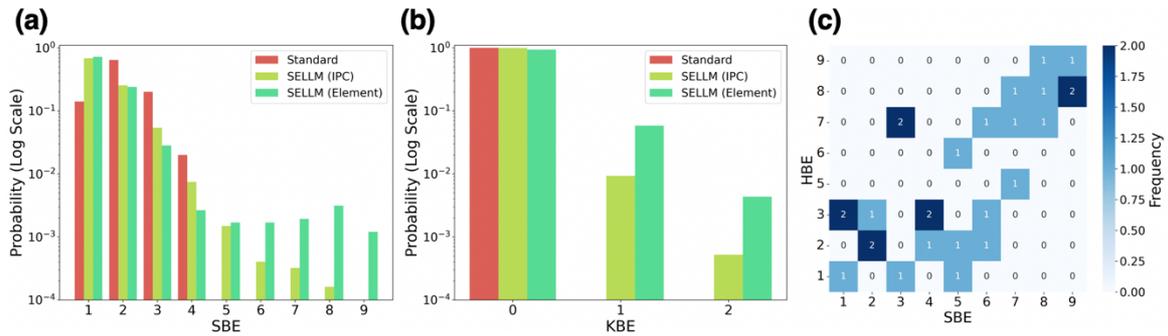

Figure 4. Evaluation of solutions generated by SELLM for the problem of contact resistance in IGZO-TFT. (a) Distribution of SBE scores for solutions generated by SELLM and the Standard approach, presented on a logarithmic probability scale. (b) Distribution of KBE evaluation scores, also presented on a logarithmic probability scale. (c) Relationship between SBE and HBE scores for a subset of generated solutions evaluated using HBE.

Table 4. Problem statement, reference solution, and keywords for the problem of contact resistance in IGZO-TFT.

| | |
|---|---|
| Problem Statement | Thin film transistors (TFTs) based on amorphous oxide semiconductors (AOSs) have garnered considerable attention for applications in next-generation storage devices such as capacitor-less dynamic-random access memory (DRAM) and high-density DRAM technologies. Such storage devices employ complex architectures with TFTs stacked vertically to achieve high storage densities. Despite their potential, AOS TFTs suffer from contact resistance issues between AOSs and electrodes resulting in excessively high contact resistance, thereby degrading charge carrier mobility, and increasing power consumption. Moreover, vertically stacked architectures further exacerbate these issues. Many methods have been proposed to address these issues, including the deposition of a highly conductive oxide interlayer between the contacts, forming oxygen vacancies on the AOS contact surface and surface treatment with plasma. Hydrogen plays a key role in these methods, as it, when dissociated into atomic hydrogen and injected into the AOS-electrode contact area, generates charge carriers, thereby reducing contact resistance. However, these methods are energy-intensive or require multiple steps and while they effectively address the high-contact resistance of the exposed upper surface of the semiconductors, they are impractical for buried contacts within the complex nanoscale architectures of storage devices.<br>Please explain how to make the scattering layer, satisfying the above requirements. |
| Reference solution[30] | Palladium, which has a high hydrogen transport capacity and can isolate hydrogen in atomic form as a catalyst, is the best candidate as the wiring and electrode material. By exposing the embedded TFT and wiring structures to a hydrogen atmosphere, hydrogen becomes atomic hydrogen on the palladium surface of the terminals and can be transferred from the wiring embedded in the insulating film to the electrodes. Furthermore, the atomic hydrogen reaches the IGZO electrode interface, and the atomic hydrogen reduces the IGZO to obtain low contact resistance. |
| Keywords | Group1: palladium and Pd<br>Group2: catalyzing, catalyst, and catalytic |

Next, evaluation using HBE was conducted. As with the light extraction case, three solutions were selected for each SBE score from 1 to 10 among those generated by SELLM, and these were evaluated by HBE. Table S2 lists the evaluated solutions and their corresponding HBE scores. Figure 4c shows the relationship between SBE and HBE scores. Similar to the light extraction problem, a correlation between SBE and HBE scores was observed overall. As highlighted in Table 6, SELLM successfully generated highly accurate solutions with an HBE score of 9. Interestingly, there are solutions with a relatively low SBE score of 3 but a comparatively high HBE score of 7. This is particularly noteworthy because it proposes an alternative solution using platinum, which, like palladium, can incorporate hydrogen, even though it is not mentioned in the reference solution. These results demonstrate that the solutions generated by SELLM are not only highly valid but also capable of generating feasible alternatives that have not been previously documented in existing research.

Table 5. Examples of solution generated by SELLM with high SBE and KBE scores by SELLM for the problem contact resistance in IGZO-TFT.

| Sub-class | Generated solution | SBE | KBE |
|---|---|---|---|
| H01K | Implement a metal-catalyzed hydrogenation process where a thin catalytic metal layer, such as palladium, is deposited at the AOS-electrode interface. When exposed to a hydrogen-rich environment, the metal layer can dissociate molecular hydrogen into atomic hydrogen, which then diffuses into the contact area, reducing resistance. This method provides an efficient, single-step process to enhance the electrical properties of AOS TFTs, suitable for complex device architectures. The process aligns with IPC code H01K by offering a novel manufacturing technique that addresses the contact resistance challenge in next-generation storage technologies. | 9 | 2 |
| C01B | Utilizing a catalytic layer of palladium or platinum in conjunction with AOS could facilitate the dissociation of molecular hydrogen into atomic hydrogen directly at the contact interface. This catalytic process lowers the activation energy required for hydrogen integration and creates a more conductive contact by generating free carriers. The catalytic layer can be deposited using sputtering techniques, which are compatible with large-scale manufacturing. This approach is efficient in terms of both energy and resources, and it is associated with IPC code C01B 3/16, which discusses the use of catalysts in hydrogen-related processes. | 8 | 2 |
| Pd | Palladium-Enhanced Electrode Materials: Develop new electrode materials incorporating palladium to directly influence the AOS contact interface. By integrating palladium into the electrode composition, it acts as a catalyst for hydrogen absorption, reducing contact resistance substantially. This approach simplifies the device architecture by eliminating additional layers. | 9 | 2 |
| Pd | Integrate a Pd-based hydrogenation layer using atomic layer deposition (ALD) techniques. This method allows for precise control over the thickness and composition of the Pd layer, enabling its use in nanoscale architectures. The Pd layer can serve as an efficient catalyst for hydrogen dissociation, continually generating charge carriers at the AOS contact points. ALD provides the advantage of conformal coating over complex topologies, ensuring even the buried contacts benefit from reduced resistance. | 9 | 2 |

Table 6. Examples of solutions with high HBE scores or cases where SBE scores were low but HBE scores were relatively high for the problem contact resistance in IGZO-TFT.

| Sub-class | Generated solution | SBE | KBE | HBE | Comment by human expert |
|---|---|---|---|---|---|
| H01K | Implement a metal-catalyzed hydrogenation process where a thin catalytic metal layer, such as palladium, is deposited at the AOS-electrode interface. When exposed to a hydrogen-rich environment, the metal layer can dissociate molecular hydrogen into atomic hydrogen, which then diffuses into the contact area, reducing resistance. This method provides an efficient, single-step process to enhance the electrical properties of AOS TFTs, suitable for complex device architectures. The process aligns with IPC code H01K by offering a novel manufacturing technique that addresses the contact resistance challenge in next-generation storage technologies. | 9 | 2 | 8 | Placing a palladium thin film in contact with AOS replicates the original inspiration. However, the specific process has not been disclosed. |
| Pd | Utilize palladium in the development of a hybrid electrode structure that combines the benefits of Pd's hydrogen absorption with other conductive materials. This hybrid structure can optimize the balance between conductivity and durability, providing a long-term solution to contact resistance issues in high-density DRAM technologies. | 9 | 1 | 9 | Placing a palladium thin film in contact with AOS replicates the original inspiration. While specific materials are not disclosed, the idea considers combining it with conductive materials to address the conductivity issues associated with palladium itself. |
| Pt | Create a dual-layer electrode structure with Pt as the bottom layer and a traditional conductive material on top. The Pt layer directly contacts the AOS, ensuring minimal contact resistance due to its high conductivity and stability. The top layer can be a more cost-effective material, reducing overall material costs while maintaining performance. This structure can be fabricated using sequential deposition techniques, ensuring compatibility with existing manufacturing technology while leveraging Pt's benefits at the critical AOS interface. | 3 | 0 | 7 | In addition to hydrogen doping using Pt, an idea has been proposed to address the reduction of electrode resistance by laminating a low-resistivity metal with Pt. It has also been shown that the process can be efficiently carried out with continuous deposition along with Pt. However, while Pt exhibits hydrogen activation, its hydrogen storage and transport capabilities are low, leaving it unclear how hydrogen can be transported to the Pt electrode. |
| C22C | A noble metal-based alloy, specifically a platinum alloy (C22C 5/04), can be used to form a conductive interlayer between AOS and electrodes. Platinum alloys are renowned for their superior conductive properties and resistance to oxidation, which is critical in reducing contact resistance. The layer can be deposited using atomic layer | 3 | 0 | 7 | Using atomic layer deposition (ALD) as a process to apply Pt film to complex structures is a viable possibility. However, while Pt exhibits hydrogen activation properties, its hydrogen storage and transport capabilities are low, making it unclear how hydrogen |

| | | | |
|---|---|---|---|
| | deposition, a technique that ensures high precision and conformity, necessary for buried contacts in complex storage architectures. This approach not only addresses the resistance issue but also improves the overall reliability and performance of the storage device. | | | can be effectively transported to the Pt electrode. |

## 3. Discussion & Conclusion

In this study, we developed and evaluated SELLM (Solution Enumeration via comprehensive List and LLM), a framework designed to generate solutions to intrinsically challenging problems by leveraging LLMs. By employing structured lists, such as IPC subclasses and chemical elements, SELLM creates domain-specific experts capable of addressing problems that require the integration of knowledge from diverse fields. We demonstrated the effectiveness of SELLM in solving two complex challenges: light extraction in OLED lighting and contact resistance in IGZO-based TFTs. The results demonstrated that SELLM generated solutions with higher SBE and KBE scores compared to the solutions generated without specific adjustments. Furthermore, expert evaluations confirmed that SELLM produced valuable solutions, including promising alternatives not documented in the existing literature. These findings suggest that SELLM has the potential to address inherently challenging problems by integrating domain knowledge comprehensively and systematically. Moreover, the results of this study also suggest that, even without complementing LLMs with external domain-specific knowledge through methods like RAG or fine-tuning, it is possible to generate high-quality, specialized solutions by effectively utilizing the knowledge already stored within LLMs. Even if the solutions generated are not perfect, presenting a list of possible solutions can inspire human users, spark new ideas, and help refine or develop incomplete ideas further.

SELLM relates to existing research on leveraging LLMs for hypothesis generation, problem-solving, and knowledge integration but adopts a distinct approach. While previous studies have demonstrated the utility of fine-tuning[1,12,13] and RAG[8,14] to enhance domain expertise, SELLM introduces an innovative methodology by systematically applying MECE principles to construct comprehensive expert lists. Unlike previous approaches that focus on individual domains or limited interactions between expert models[6,16], SELLM generates exhaustive and diverse solutions that connect seemingly unrelated fields. This study complements and extends recent advancements in LLM-based problem-solving[18,20] by addressing the limitations in generating solutions for intrinsically challenging problems that require the integration of distant knowledge.

From the perspective of hypothesis construction, this study is closely aligned with Charles Peirce's concept of abduction, also referred to as retroduction[44–46]. Abduction, as defined by Peirce, is the process of forming plausible hypotheses to explain or address a phenomenon. Our results demonstrate that, by providing structured guidance, SELLM can generate a multitude of hypotheses, including useful solutions, which align with Peirce's framework of abduction. Nevertheless, as Peirce emphasized, abduction involves not only the generation of hypotheses but also their refinement to identify the most valuable ones from a broad pool of candidates. While SELLM succeeds in generating a comprehensive set of hypotheses, it does not yet adequately address the critical process of narrowing these down to the most beneficial solutions.

Closely related to the previous discussion, one limitation of SELLM is its tendency to generate redundant or irrelevant ideas due to the exhaustive nature of its approach. While this redundancy may be inherent to the process of fostering innovative ideas, it poses a substantial obstacle in practical applications. Currently, SELLM lacks built-in mechanisms to filter out less viable ideas, leaving this task to human experts or subsequent analytical tools. Future research should focus on developing efficient methods to evaluate and rank the generated solutions, reducing noise while preserving the breadth of innovation.

A key advantage of SELLM lies in its ability to control solution generation by selecting appropriate lists, such as company-specific technologies or laboratory resources. This flexibility enables tailored applications across industries and research domains. For instance, SELLM could be used to identify potential uses for proprietary materials, suggest applications for underutilized technologies, or explore new combinations of available resources. In addition, integrating advanced filtering mechanisms and interactive feedback loops with human experts could further enhance SELLM's utility.

## 4. Methods
### 4.1 Lists of specific knowledge and their explanations

SELLM generates a list of experts from a list of terms consisting of concepts and domain knowledge and outputs solutions using the expert list. The list of knowledge and concepts must be sufficiently comprehensive and appropriately granular to include concepts that contribute to the solution generation. Additionally, excessive overlap between concepts in a list can lead to redundant and similar responses. Therefore, a MECE (Mutually Exclusive, Collectively Exhaustive) list is desirable whenever possible.

In this study, we utilized a list of subclasses from the International Patent Classification (IPC) and a list of chemical elements. The IPC is a representative system for structurally organizing technologies across various fields, providing a sufficiently comprehensive and non-overlapping list of technologies. The patent classification consists of sections A through H. While it is possible to use all sections and subclasses, we selected sections B (performing operations; transporting), C (chemistry; metallurgy), F (mechanical engineering; lighting; heating; weapons; blasting), G (physics), and H (electricity) to consider computational cost. The numbers of subclasses in sections B, C, F, G, and H are 170, 87, 99, 87, and 54, respectively, resulting in a total of 497 subclasses. For the list of chemical elements, we used the 83 element symbols from hydrogen (element number 1) to bismuth (element number 83), which are deemed practically applicable under standard conditions.

To accurately generate experts that reflect the concepts included in a list, descriptive texts were prepared for each concept. For IPC, descriptions were generated based on the subclass symbols, including the subclass titles and references, as well as the symbols and titles of the main groups and subgroups to which each subclass belongs, thereby explaining the technologies associated with each subclass. For the elements, descriptive texts were generated for each individual element. The generation of these descriptive texts was performed using GPT-4o. The prompts used for creating these descriptions are presented in Table S3.

### 4.2 Generation of Experts and Solutions

From the comprehensive list prepared, domain-specific expert LLMs were systematically created for each technological field. Specifically, expert LLMs were generated by instructing the solution-generating LLM to adopt expert roles through role-play prompting, as outlined in Table S4. Each expert LLM was tasked with generating 10 solutions. To ensure stable evaluation, the creation of experts and the generation of solutions were repeated five times.

Due to the large number of experts required for solution generation, the process incurred significant computational cost and time. To address this, in the cases of light extraction challenges and IGZO-TFT issues, the relatively cost-efficient and faster GPT-4o model was used for both expert creation and solution generation. GPT-4o was accessed via its API with default parameters, including a temperature setting of 0.7.

### 4.3 Evaluations for Generated Solutions

The generated solutions were evaluated using three metrics: SBE, KBE, and HBE. The evaluation of SBE was performed by comparing the generated solutions to reference descriptions of solutions for the light extraction and IGZO-TFT challenges, as provided in Table 1 and 2. The similarity was assessed based on the LLM-as-a-judge prompting, detailed in Table S5. GPT-4o was used to evaluate the similarity scores throughout this study. To improve scoring accuracy, 20 sample solutions were pre-generated for each task and evaluated by human experts. The scores and reasons assigned by the experts were used as reference data during the evaluation process. The scoring for the light extraction problem are shown in Table S6, while those for the problem of contact resistance in IGZO-TFT are provided in Table S7.

KBE focused on the presence of keywords deemed critical for addressing the challenges. The list of important keywords for each challenge is provided in Table 1 and 2. For evaluation, two groups of keywords were defined for each challenge. The generated solutions were checked to determine whether they contained at least one of the keywords from either group. A KBE score of 1 or 2 indicates that the solution contains keywords from one group or from both groups, respectively.

For HBE, three solutions were extracted for each value range of the SBE metric. These solutions were then evaluated by some authors of this paper, who are materials science experts. The evaluators provided scores and justification for each solution (Tables S1 and S2). Importantly, the scoring criteria for HBE did not based on similarity to the reference solutions used in SBE. Instead, the focus was on whether the solution could address the challenge by leveraging knowledge from different fields. Consequently, solutions with high HBE scores may include innovative, yet unreported, and effective alternative approaches to solving the challenges.

**Data and code availability**

The code of SELLM is available on GitHub at https://github.com/ycu-iil/SELLM.

**Acknowledgement**

This work was supported by the Ministry of Education, Culture, Sports, Science and Technology (MEXT) under the grants: Data Creation and Utilization Type Material Research and Development Project (Grant Number: JPMXP1122683430), Simulation and AI-driven next-generation medicine and drug discovery based on "Fugaku" (Grant Number: JPMXP1020230120), and Feasibility studies for the next-generation computing infrastructure. This research was also supported by the JST FOREST JPMJFR232U.

**Competing interests statement**

The authors declare no competing interests.

**Contribution**

K.T. and N.N. designed the study. H.T. performed programming, collecting the dataset, and analyzing the result. N.N. and T.K. analyzed the generated solutions. K.T. and H.T. wrote the manuscript. S.I. checked the program, manuscript, and data. All the authors reviewed and approved the final manuscript.

**Supplemental Information for**

Extracting effective solutions hidden in large language models via generated comprehensive specialists: case studies in developing electronic devices

**Author:**

Hikari Tomita[1,#], Nobuhiro Nakamura[2,3,#], Shoichi Ishida[4], Toshio Kamiya[2], Kei Terayama[1,2,4,*]

**Affiliation:**

[1]School of Science, Yokohama City University, Tsurumi-ku, Yokohama 230-0045, Japan

[2]MDX Research Center for Element Strategy, Institute of Science Tokyo, Yokohama, 226-8501, Kanagawa, Japan.

[3]Materials Integration Laboratories, AGC Inc, Yokohama, 230-0045, Japan

[4]Graduate School of Medical Life Science, Yokohama City University, Tsurumi-ku, Yokohama 230-0045, Japan

#equally contributed

*Corresponding author(s). E-mail(s): terayama@yokohama-cu.ac.jp

Table S1. Examples of solutions generated by SELLM for the light extraction problem, with their HBE scores and comments from human experts.

| Sub-class | Generated solution | SBE | KBE | HBE | Comment by human expert |
|---|---|---|---|---|---|
| C03C | Formulate a scattering layer using a high-refractive-index glass frit. The frit, consisting of finely ground glass, is mixed with a binding agent to form a paste, which is then spread on a substrate and fired to form a dense, uniform layer. The glass composition should include high-index elements such as lead or bismuth oxides to enhance refractive properties. Post-firing, the layer is polished to achieve the required smoothness. This solution is in line with IPC code C03C 6/00, which focuses on glass batch compositions and frit technologies for modifying optical characteristics. | 9 | 2 | 7 | The idea of using a high-refractive-index glass frit paste to form a scattering layer during the firing process replicates the original solution but includes an unnecessary polishing process. There is a lack of understanding that a smooth surface can be achieved through the firing process. While there is no mention of scattering particles, the bubbles present in the fired layer act as substitutes for scattering particles, so there is no issue. |
| G03F | Explore the use of a high-refractive-index glass frit sintering process to form a scattering layer. By selecting a glass composition with a refractive index of 1.9 or higher, the frit can be sintered onto the glass substrate, forming a smooth and uniform layer. The process involves heating the frit to a temperature where it becomes viscous and flows to form a flat layer. This method ensures a seamless interface with the substrate, reducing light loss. The sintering process is aligned with the IPC code G03F, which includes processes for creating textured surfaces. | 9 | 2 | 8 | The idea of using a high-refractive-index glass frit paste to form a scattering layer during the firing process replicates the original solution. Although there is no mention of scattering particles, the bubbles present in the fired film serve as substitutes for scattering particles, so there is no issue. |
| C03B | A scattering layer can be manufactured by using a two-step thermal treatment process. First, a high refractive index glass frit, such as barium oxide-based glass, is prepared by melting the raw materials, as detailed in IPC code C03B 5/00, and then ground into fine particles. These particles are mixed with a binder to form a paste. The paste is applied to a glass substrate and initially heated at a lower temperature to evaporate the binder and allow the particles to adhere. A secondary higher temperature treatment follows, which melts and fuses the particles into a continuous layer. This process allows control over the layer's thickness and surface smoothness, preventing short circuits in OLEDs. | 9 | 1 | 9 | The idea of using a high-refractive-index glass frit paste to form a scattering layer during the firing process replicates the original solution. It also mentions glass materials for achieving a high refractive index. Although there is no mention of scattering particles, the bubbles present in the fired film act as substitutes for scattering particles, so there is no issue. |
| C23D | Design a vitreous scattering layer specifically for OLEDs by tailoring the enamel composition with barium or strontium oxides, which are known to increase the refractive | 8 | 2 | 9 | The idea of using a high-refractive-index glass frit paste to form a scattering layer during the firing process replicates the |

| | index. The enamel is applied using a wet method (C23D 5/02) to maintain surface smoothness, essential for preventing electrical shorts. After application, a controlled firing process (C23D 9/00) ensures the enamel is clear and smooth, optimizing light extraction from the OLED. | | | | original solution. It also mentions glass materials for achieving a high refractive index. Although there is no mention of scattering particles, the bubbles present in the fired film act as substitutes for scattering particles, so there is no issue. |
|---|---|---|---|---|---|
| H01M | Develop a composite material using a high refractive index glass frit mixed with a low melting point glass to form a smooth scattering layer. The glass frits, such as lead or barium-based glasses, exhibit high refractive indices and can be sintered onto a substrate at relatively low temperatures. This process ensures a smooth, homogenous layer that enhances light extraction without causing short circuits. The approach involves mixing, pressing, and sintering techniques akin to those in electrode manufacturing, aligning with IPC H01M's focus on material processing and formation methods. | 8 | 2 | 8 | The idea of firing high-refractive-index glass frit with low-melting-point glass to obtain a scattering layer is a good one. It is also possible for the low-melting-point glass to serve the scattering function. However, if the content of low-melting-point glass is too high, it may prevent the formation of a high-refractive-index matrix, which could result in cases where the improvement in light extraction efficiency is not achieved. |
| B32B | Utilizing B32B 17/00, a glass-based approach can be taken to develop the scattering layer. Glass with naturally high refractive indices, such as lead glass, can be used as the base material. To enhance its light scattering properties, the glass can be embedded with micron-sized air voids or bubbles during the manufacturing process, creating a microstructured surface. This structure can scatter light efficiently while maintaining an overall smooth surface due to the fine control in glass fabrication. Additionally, the refractive index of the glass can be further tuned by doping with metal oxides. This method leverages traditional glass-making techniques while introducing innovative scattering mechanisms to improve light extraction in OLEDs. | 8 | 0 | 7 | The idea of applying the substrate itself high-refractive-index is theoretically feasible. Scattering particles are also mentioned. However, from an industrial perspective, it would be challenging to manufacture high-refractive-index glass into a plate shape that includes bubbles while maintaining a smooth surface. |
| F27B | For a cost-effective solution, consider using open sintering apparatus (F27B 21/00) to produce a high-refractive-index glass frit that can be melted and cast into a scattering layer. This apparatus allows for large-scale production and flexibility in adjusting sintering parameters to optimize refractive index and mechanical properties. After casting, the layer is polished to achieve the necessary smoothness, ensuring it meets the stringent requirements for OLED light extraction efficiency and electrical safety. | 7 | 2 | 6 | It is difficult to cast high-refractive-index glass frit to form a scattering layer. |
| B26B | Creating a scattering layer using a high-refractive-index ceramic material offers a robust solution. Materials like barium titanate (BaTiO3) or lead zirconate titanate (PZT) can be processed into thin films using techniques such as tape casting or screen printing. These ceramics inherently have high refractive indices and can be sintered to achieve a dense, smooth surface. The sintering process can be optimized to prevent surface roughness, ensuring the layer's suitability for OLED applications. The durability and thermal stability of ceramic materials make them ideal for long-term use in devices where environmental stability is crucial. This solution effectively combines high refractive index with the necessary surface quality for efficient light extraction. | 7 | 1 | 5 | It is possible to obtain a high-refractive-index scattering layer by firing ceramic particles; however, their inability to soften and flow makes it difficult to achieve a smooth surface. |
| G02F | A ceramic scattering layer can be produced using materials such as lead zirconate titanate (PZT) or barium titanate, which have high refractive indices and are discussed under IPC G02F 1/03. These ceramics can be sintered to form a dense, smooth layer that reduces the risk of electrical shorts in the OLED. The layer can be fabricated by screen-printing a ceramic slurry onto the glass substrate, followed by controlled drying and sintering. The process allows for the production of a scattering layer with a precisely defined thickness and surface smoothness. By optimizing the sintering conditions, the refractive index and optical scattering properties can be tailored to maximize light extraction from the OLED. This method leverages the inherent properties of electro-optical ceramics to enhance device performance. | 7 | 1 | 5 | It is possible to obtain a high-refractive-index scattering layer by firing ceramic particles; however, their inability to soften and flow makes it difficult to achieve a smooth surface. Additionally, when sintering ceramics on a glass substrate, the high sintering temperature may cause the glass to deform. |
| H05B | Develop a hybrid scattering layer using a combination of inorganic and organic materials. Utilize a high-refractive-index glass, such as lead glass, embedded within a polymer matrix. The glass particles (refractive index > 1.9) facilitate scattering, while the polymer ensures a smooth surface. Use hot-pressing or extrusion methods to form the layer, achieving uniform thickness and surface smoothness. This hybrid approach balances the refractive index and mechanical properties, aligned with IPC H05B's focus on enhancing electroluminescent device performance through material innovation. | 6 | 0 | 3 | Polymers are used as the matrix for scattering layers, but with their normal refractive indices, they do not improve light extraction efficiency. |
| C22B | A scattering layer with a high refractive index can be fabricated using a layer of lead glass, known for its high refractive index and optical | 6 | 0 | 3 | The idea of making the substrate itself high-refractive-index is feasible, but since it |
| | transparency. The lead glass can be prepared by refining lead-containing ores or scrap using techniques under the C22B subclass, such as chloridising or sulfating roasting, which help to purify the lead before it is incorporated into the glass. The glass can then be melted and cast into thin sheets. These sheets are polished to achieve a smooth surface suitable for OLED application. The high refractive index of lead glass (typically greater than 1.9) makes it an excellent choice for enhancing light extraction from OLEDs while ensuring the surface remains smooth. | | | | lacks scattering properties, it does not improve light extraction efficiency. |
| B24D | To achieve a high-quality scattering layer, an inorganic bonding approach using close-grained structures as per B24D 3/04 can be employed. This involves using materials like barium titanate or strontium titanate, which provide high refractive indices. The process includes mixing these particles with a suitable inorganic binder, spreading the mixture over the glass substrate, and subsequently firing it to form a dense, smooth layer. The close-grained nature of the resulting material ensures minimal porosity, thereby enhancing the refractive index and providing a robust surface that prevents electrode shorting. This technique is particularly suitable for applications requiring high thermal stability and mechanical strength, making it ideal for OLED devices operating under varying environmental conditions. | 6 | 1 | 5 | It is possible to obtain a high-refractive-index scattering layer by firing ceramic particles; however, their lack of softening and flow makes it difficult to achieve a smooth surface. Additionally, sintering ceramics on a glass substrate may pose a risk of glass deformation due to the excessively high sintering temperature. |
| B21J | Another approach under B21J 9/00 involves using forging presses to shape a high-refractive-index glass layer. The glass can be initially heated to a malleable state and then forged into thin sheets. This process allows precise control over the thickness and smoothness of the glass. By selecting a glass composition with a refractive index above 1.9, such as lead glass, the scattering layer can be optimized for light extraction. The smooth surface is ensured by polishing post-forging, resulting in a layer suitable for OLED fabrication. | 5 | 0 | 3 | The idea of making the substrate itself high-refractive-index is feasible, but since it lacks scattering properties, it does not improve light extraction efficiency. |
| G04B | Inspired by the protective measures in mechanically-driven clocks, a multi-layered scattering layer can be designed. The core layer can be a high refractive index glass or polymer composite, while a protective, smooth outer layer prevents electrical shorts. This mirrors the use of protective cases in G04B 37/00, where clocks are protected against shocks and environmental factors. The high refractive | 5 | 0 | 1 | The description is unclear or lacks sufficient information to make a judgment. |
| | index layer can be made from a sol-gel process, embedding high-index materials like hafnium oxide (HfO2) within a silica matrix. Post-deposition polishing ensures a smooth surface. This method ensures that the refractive index requirements are met without compromising the mechanical stability of the OLED structure, similar to how mechanical clocks ensure durability and precision through meticulous case design and material selection. | | | | |
| C04B | A scattering layer made from a mixture of high-refractive-index ceramic powders, such as yttrium oxide (Y2O3), and a binder can be created. This mixture is applied to a substrate using techniques like screen printing or doctor blading, followed by sintering to enhance the refractive index while ensuring a smooth surface. The use of yttrium oxide aligns with the requirements for high refractive index materials and is supported by the guidelines of C04B 35/00, which focus on ceramic compositions and their processing methods to achieve desired optical and mechanical properties. | 5 | 1 | 5 | It is possible to obtain a high-refractive-index scattering layer by firing ceramic particles; however, their inability to soften and flow makes it difficult to achieve a smooth surface. The binder is presumed to be inorganic, but achieving both smoothness and enhanced light extraction is highly challenging. |
| B33Y | Implement a multi-material 3D printing approach (IPC code B33Y 70/00) to produce a composite scattering layer. Use a high-refractive-index polymer as the base material and incorporate ceramic particles like zirconia (ZrO2) for enhanced refractive properties. This technique allows for precise layer-by-layer deposition, ensuring uniformity and smoothness across the layer's surface. Post-printing polishing can be employed to achieve the necessary smoothness for OLED integration. | 4 | 0 | 1 | The description is unclear or lacks sufficient information to make a judgment. |
| H01T | Inspired by H01T's approach to managing electrical discharges, consider fabricating the scattering layer through layer-by-layer (LbL) assembly of high-refractive-index materials such as zinc oxide or titania. This method allows precise control over film thickness and composition, crucial for achieving the desired refractive index and smooth surface. Begin by preparing a substrate with alternating polyelectrolyte layers interspersed with nanoparticle layers, ensuring each layer is ultra-thin and uniform. The LbL technique naturally produces smooth surfaces, reducing the risk of short circuits in OLEDs. Post-assembly, the film can be sintered at low temperatures to enhance structural integrity without compromising smoothness. This method reflects the precision and control | 4 | 1 | 1 | The description is unclear or lacks sufficient information to make a judgment. |
| | characteristic of H01T, which is essential for tailoring electrical properties in spark gaps and ion generators. | | | | |
| H04J | Adapting the concept of orthogonal multiplexing, a scattering layer with anisotropic properties can be developed for OLEDs. By aligning elongated nanoparticles such as ZnO nanorods within a polymer matrix, the layer can scatter light differently depending on its polarization. This selective scattering enhances light extraction while maintaining a smooth surface. The alignment is achieved through techniques like electrospinning, where the electric field controls nanoparticle orientation. The approach is akin to managing orthogonal channels in communication systems by using material anisotropy. | 4 | 0 | 2 | It is difficult to obtain the described scattering layer using electrospinning. Polarization control is unnecessary, and with ordinary polymers, light extraction efficiency cannot be improved. |
| F24B | 9. Utilize a bio-inspired approach by creating a scattering layer mimicking the structure of butterfly wings, which naturally have high refractive indices and smooth surfaces. This can be achieved through a combination of soft lithography and biomimetic deposition techniques. The resulting layer is both high-index and smooth, suitable for OLED applications. This method reflects F24B's use of natural design principles in improving efficiency and functionality of heating devices. | 3 | 0 | 1 | The description is unclear or lacks sufficient information to make a judgment. |
| H10N | A novel approach involves embedding high-index quantum dots within a transparent polymer matrix to create a scattering layer for OLEDs. Quantum dots like cadmium selenide (CdSe) or lead sulfide (PbS) can be engineered to exhibit refractive indices above 1.9, while the polymer matrix ensures a smooth, defect-free surface. Spin-coating or inkjet printing can be utilized to apply this composite layer, ensuring uniformity and scalability. The quantum dots not only enhance the refractive index but also have the potential to modify the emission spectrum, offering additional customization for OLED applications. This solution reflects the cutting-edge advancements in nanotechnology and solid-state device engineering under H10N, demonstrating the potential for novel material applications in enhancing device performance. By integrating quantum dot technology, this method supports the development of highly efficient and customizable OLED systems. | 3 | 0 | 3 | Quantum dots do not scatter visible light. |
| B28C | Construct a scattering layer using a high-index clay mixed with reflective metallic particles, such as silver or aluminum flakes. These | 3 | 0 | 1 | It is impossible to fabricate OLEDs on clay. |
| | particles enhance light scattering due to their reflective properties. Use mechanical means (B28C 5/08) to mix the clay with metallic flakes, ensuring an even distribution of particles. This process helps achieve a composite material with a high refractive index and effective scattering properties. Apply the mixture onto a substrate using a smooth applicator to maintain a flat surface, crucial for OLED performance. Control the particle size and distribution to optimize scattering without increasing surface roughness. This method combines the reflective properties of metallic particles with the stability of clay, creating an efficient light-extracting layer with a smooth finish. | | | | |
| G02C | 6. **Incorporating Micro-lens Arrays:** Design a scattering layer with embedded micro-lens arrays to improve light extraction, drawing parallels with G02C 7/10, which deals with optical enhancements in lenses. These arrays can redirect trapped light towards the substrate-air interface, increasing efficiency. The smoothness of the layer can be controlled through precise fabrication techniques inherent in lens technology. | 2 | 0 | 3 | Forming a microlens array with scattering properties does not improve light extraction efficiency unless the refractive index is high. While a high refractive index can potentially enhance light extraction efficiency, using high-refractive-index resins presents challenges such as coloration and reliability concerns. |
| G16C | Under IPC G16C 20/50, molecular design tools can be used to create novel polymers with inherently high refractive indices. By manipulating the polymer backbone and side chains, one can design polymers that not only meet the refractive index criteria but also exhibit properties conducive to forming smooth films. Computational tools can predict the effect of different structural modifications on the refractive index and mechanical properties of the polymers, enabling the design of materials that are both effective as scattering layers and easy to process, thus preventing short-circuiting in OLEDs. | 2 | 0 | 3 | Moisture within high-refractive-index polymer films is likely to cause damage to the OLED device. |
| B22F | One approach to creating a scattering layer with a high refractive index is to use a metallic powder with a refractive index equal to or greater than 1.9. This powder can be mixed with a transparent polymer matrix to form a composite material. The metallic powder can be selected based on its optical properties and compatibility with the polymer. Techniques like ball milling or ultrasonic dispersion can be used to achieve a uniform distribution of the powder within the matrix. The composite can be molded into thin films using a process such as hot pressing or extrusion, ensuring a smooth | 2 | 0 | 3 | Metals have absorption properties, causing light attenuation and preventing improvement in light extraction efficiency. |
| | surface to prevent short circuits. The smoothness can be enhanced by polishing or using a mold with a high-quality finish. This approach falls under IPC B22F, focusing on the use of metallic powders in composite materials. | | | | |
| F01D | Inspired by the blade cooling techniques in F01D 5/08, a scattering layer can be designed to include microchannels filled with a refractive fluid. These microchannels simulate the cooling channels in turbine blades, but instead, serve as light-guiding paths. The refractive fluid can be a high-index liquid, such as an oil-based or polymeric solution, ensuring the overall refractive index of the layer is maintained at or above 1.9. The solid matrix of the scattering layer can be made from silica or another transparent material, polished to a smooth finish to prevent electrode short-circuiting. This design not only increases light extraction but also allows for dynamic tuning of the refractive properties by altering the fluid composition. | 1 | 0 | 1 | The description is unclear or lacks sufficient information to make a judgment. |
| G06T | 3D image generation techniques under G06T 17/00 can assist in visualizing the interaction of light with the scattering layer. By creating a detailed 3D model of the scattering layer, the flow of light can be simulated to identify areas where light extraction can be improved. This visualization helps in refining the scattering layer's design, ensuring it meets the required refractive index and smoothness criteria. | 1 | 0 | 1 | The description is unclear or lacks sufficient information to make a judgment. |
| B23K | 9. Explosive welding techniques can be adapted to create a smooth high-index layer by explosively bonding a thin film of TiO2 onto the glass substrate. This method uses controlled explosions to apply pressure and heat, resulting in a smooth and uniform layer. Though primarily used for metals, this technique can be modified for non-metallic, high-refractive-index materials, ensuring a smooth finish needed for OLEDs. Explosive welding is categorized under non-electric welding methods in B23K. (IPC: B23K 20/00) | 1 | 0 | 2 | It is challenging to fabricate a scattering layer. |

Table S2. Examples of solutions generated by SELLM for problem contact resistance in IGZO-TFT and their HBE scores and comments from human experts.

| Sub-class or Element | Generated solution | SBE | KBE | HBE | Comment by human expert |
|---|---|---|---|---|---|
| H01K | Implement a metal-catalyzed hydrogenation process where a thin catalytic metal layer, such as palladium, is deposited at the AOS-electrode interface. When exposed to a hydrogen-rich environment, the metal layer can dissociate molecular hydrogen into atomic hydrogen, which then diffuses into the contact area, reducing resistance. This method provides an efficient, single-step process to enhance the electrical properties of AOS TFTs, suitable for complex device architectures. The process aligns with IPC code H01K by offering a novel manufacturing technique that addresses the contact resistance challenge in next-generation storage technologies. | 9 | 2 | 8 | Placing a palladium thin film in contact with AOS replicates the original solution. However, the specific process has not been disclosed. |
| Pd | 7. **Palladium-Enhanced Electrode Materials**: Develop new electrode materials incorporating palladium to directly influence the AOS contact interface. By integrating palladium into the electrode composition, it acts as a catalyst for hydrogen absorption, reducing contact resistance substantially. This approach simplifies the device architecture by eliminating additional layers. | 9 | 2 | 8 | Placing a palladium thin film in contact with AOS replicates the original solution. However, the meaning of "no additional layer required" is unclear. |
| Pd | 10. Utilize palladium in the development of a hybrid electrode structure that combines the benefits of Pd's hydrogen absorption with other conductive materials. This hybrid structure can optimize the balance between conductivity and durability, providing a long-term solution to contact resistance issues in high-density DRAM technologies. | 9 | 1 | 9 | Placing a palladium thin film in contact with AOS replicates the original solution. While specific materials are not disclosed, the idea considers combining it with conductive materials to address the conductivity issues associated with palladium itself. |
| B01J | Develop a catalytic process to enhance the dissociation of hydrogen molecules into atomic hydrogen at the AOS-electrode interface. This can be achieved by integrating a catalyst layer, such as palladium or platinum, which facilitates hydrogen dissociation under mild conditions. The catalytic layer can be deposited using methods described in B01J 8/00, focusing on chemical processes with solid catalysts. This approach allows atomic hydrogen to penetrate deeper into the contact | 8 | 2 | 7 | Using palladium is a good approach, but the process described in B01J 8/00 is a wet process and cannot be applied to the current issue. |
| | region, significantly reducing contact resistance. The use of catalysts ensures that the process is efficient and minimizes energy consumption, making it suitable for application in buried contact areas within vertically stacked architectures. The integration of such catalytic systems can be seamlessly incorporated into existing fabrication processes, offering a practical solution to the contact resistance challenge. | | | | |
| C01B | Utilizing a catalytic layer of palladium or platinum in conjunction with AOS could facilitate the dissociation of molecular hydrogen into atomic hydrogen directly at the contact interface. This catalytic process lowers the activation energy required for hydrogen integration and creates a more conductive contact by generating free carriers. The catalytic layer can be deposited using sputtering techniques, which are compatible with large-scale manufacturing. This approach is efficient in terms of both energy and resources, and it is associated with IPC code C01B 3/16, which discusses the use of catalysts in hydrogen-related processes. | 8 | 2 | 8 | Placing a palladium catalyst layer in contact with AOS replicates the original solution. Pt may also function in a similar manner. |
| Pd | Design a Pd-based multilayer structure to serve as an electrode in AOS TFTs. This structure can consist of alternating layers of palladium and another conductive material, creating a composite with superior electrical and thermal properties. The palladium layers act as hydrogen reservoirs, aiding in contact resistance reduction, while the alternating material can be optimized for mechanical strength and further conductivity. Such a multilayer approach can address issues related to both contact resistance and mechanical stability in vertically stacked architectures, offering a robust solution for high-density storage devices. | 8 | 1 | 9 | Placing a palladium thin film in contact with the AOS replicates the original solution. While specific materials are not disclosed, the idea considers the conductivity issues inherent to pure palladium by combining it with a conductive material. |
| H01H | Inspired by the use of noble metals in H01H 1/02, a thin interfacial layer of an alloy containing hydrogen-storing metals such as palladium or titanium can be deposited. These metals not only reduce contact resistance but also facilitate the diffusion and stabilization of hydrogen at the contact interface. The alloy layer acts as a conductive bridge, enhancing electron mobility and reducing overall device power consumption, suitable for both exposed and buried contacts. | 7 | 1 | 7 | The palladium contact layer functions effectively, but titanium does not activate hydrogen and therefore does not function in this context. |
| C01G | Adopting a palladium oxide (C01G 5/02) layer can provide a solution for high contact resistance in AOS TFTs. Palladium oxide is | 7 | 2 | 5 | Palladium oxide can act as a hydrogen catalyst; however, it lacks hydrogen storage and |
| | known for its exceptional electrical conductivity and catalytic properties, making it an ideal choice for contact layers. The deposition of palladium oxide can be achieved through electron-beam evaporation, allowing precise control over the thickness and uniformity of the layer. This method also facilitates the incorporation of hydrogen at the contact interface, enhancing charge carrier generation and reducing resistance. Furthermore, palladium oxide's stability under various environmental conditions ensures long-term device reliability, which is essential for applications in high-density DRAM technologies. | | | | transport capabilities, which may prevent it from supplying hydrogen to the AOS. |
| Pd | Explore the use of palladium in forming a gradient layer between the AOS and electrodes. This gradient layer can be engineered to gradually change composition from pure AOS to pure Pd, facilitating smooth transition and improved adherence at the contact interface. Such a layer can be deposited using a pulsed laser deposition technique, enabling precise control over the gradient profile. This innovation can significantly reduce contact resistance and enhance the mechanical stability of the TFT stack. | 7 | 1 | 8 | Placing a palladium thin film in contact with AOS replicates the original solution. However, it is unclear whether a palladium concentration gradient contributes to improved the contact properties. |
| C22C | Explore the use of noble metal-based alloys, such as a platinum group metal alloy (C22C 5/04), to create a highly conductive interlayer. These alloys can be tailored to include elements like palladium or rhodium, enhancing their compatibility with AOS materials. Using techniques like thin film deposition via sputtering or evaporation, the noble metal alloy can form an effective bridge between AOS and electrodes, minimizing contact resistance. Despite higher material costs, the superior electrical properties and stability of noble metal alloys justify their use in critical applications where performance is paramount. | 6 | 1 | 7 | Palladium is disclosed, but the reason for its use is unclear. Rhodium has hydrogen activation functionality but lacks hydrogen storage and transport capabilities. |
| B60C | Introduce a gas-permeable membrane technology inspired by tyre valve arrangements (B60C 29/00) to inject atomic hydrogen directly into the AOS-electrode interface. By designing a membrane that selectively allows hydrogen permeation, we can maintain a continuous supply of hydrogen atoms to reduce contact resistance effectively. This approach eliminates the need for complex and energy-intensive surface treatments by facilitating a constant, | 6 | 0 | 2 | The application of a hydrogen permeable membrane is not feasible. |
| | controlled introduction of hydrogen at the contact site, ensuring low resistance even in buried interfaces within vertically stacked architectures. | | | | |
| Pd | 3. Implement Pd nanoparticles dispersed within the AOS matrix to act as localized hydrogen sources. These nanoparticles can continuously supply hydrogen to the contact regions, maintaining low resistance even in complex, multi-layered architectures. Utilizing palladium in this dispersed form can improve the uniformity of hydrogen distribution and minimize contact resistance. | 6 | 1 | 3 | Since hydrogen cannot be supplied to palladium nanoparticles dispersed within the AOS film, no effect can be expected. |
| B27K | Drawing from B27K 5/00's focus on chemical treatments, a novel chemical treatment process can be developed where a hydrogen-rich solution is used to treat the AOS contact surfaces. This solution can be engineered to penetrate deeply into the contact areas, even those buried within complex architectures. Once applied, a mild activation via heat or UV light can trigger the release of atomic hydrogen, enhancing charge carrier generation and reducing contact resistance. This method offers a significant reduction in processing steps and energy consumption compared to traditional plasma-based methods, making it well-suited for the efficient production of next-generation DRAM devices. | 5 | 0 | 2 | The wet process described is difficult to apply to ultra-fine structures. |
| F23G | Apply F23G 7/07's catalytic combustion processes to develop a low-energy method for generating reactive species capable of modifying AOS surfaces. The catalytic process can be tailored to produce a controlled flow of hydrogen atoms or other reactive species, which interact with the AOS surface to create charge carriers and reduce contact resistance. This method is particularly suitable for buried contacts, providing a uniform treatment across all interfaces. | 5 | 1 | 1 | The description is unclear or lacks sufficient information to make a judgment. |
| Ti | Employ titanium hydride (TiH2) as a hydrogen donor for in-situ hydrogenation within AOS TFTs. TiH2 can release hydrogen at controlled rates, allowing for the precise introduction of atomic hydrogen into the contact areas without external energy input. This method simplifies the hydrogenation process, aligning with energy-efficient manufacturing goals while effectively reducing contact resistance and enhancing carrier mobility within the device. | 5 | 0 | 6 | If a $TiH_2$ thin film can be deposited by sputtering, it could release hydrogen during subsequent heat treatment processes, making it a potentially viable idea. |
| G21B | 3. Utilize particle injectors (G21B 1/15) in the fabrication process to introduce atomic | 4 | 0 | 2 | The process described is difficult to apply to ultra-fine |
| | hydrogen precisely at the AOS-electrode interface. This approach allows targeted delivery of hydrogen, minimizing energy use and ensuring efficient charge carrier generation. The precise control of hydrogen injection aids in reducing contact resistance without affecting the buried contacts adversely. | | | | structures. |
| In | Introduce atomic hydrogen through indium's ability to release hydrogen ions at lower energy levels. By embedding an indium-based compound within the AOS, atomic hydrogen can be generated in situ, reducing energy requirements. This will improve the contact resistance without the need for external hydrogen sources, making the process more efficient and applicable to buried contacts in complex storage architectures. | 4 | 0 | 3 | It is considered that there are no indium compounds with excellent hydrogen storage capabilities. |
| Ti | Incorporate titanium nitride (TiN) as an electrode material due to its excellent conductivity and stability. TiN can be deposited using physical vapor deposition (PVD) techniques to form a robust and conductive layer that minimizes contact resistance. Its compatibility with AOS materials and resistance to chemical degradation make it ideal for vertically stacked TFT architectures. TiN's inherent robustness can withstand the thermal and mechanical stresses of multi-layered structures, providing a stable contact interface. | 4 | 0 | 3 | There seems to be no clear reason why TiN thin film would reduce contact resistance. |
| C09J | Developing adhesives that incorporate hydrogen-donating compounds can effectively reduce contact resistance in AOS TFTs. These compounds, integrated within an adhesive matrix, can release hydrogen atoms when subjected to mild thermal activation or catalytic action (C09J 5/00). The released hydrogen atoms help in passivating defects and enhancing charge carrier density at the AOS-electrode interface. The adhesive application ensures uniform distribution and adhesion across complex device geometries, addressing the challenge of buried contacts. | 3 | 1 | 1 | The description is unclear or lacks sufficient information to make a judgment. |
| Pt | Create a dual-layer electrode structure with Pt as the bottom layer and a traditional conductive material on top. The Pt layer directly contacts the AOS, ensuring minimal contact resistance due to its high conductivity and stability. The top layer can be a more cost-effective material, reducing overall material costs while maintaining performance. This structure can be fabricated using sequential | 3 | 0 | 7 | In addition to hydrogen doping using Pt, an idea has been proposed to address the reduction of electrode resistance by laminating a low-resistivity metal with Pt. It has also been shown that the process can be efficiently carried out with continuous |
| | deposition techniques, ensuring compatibility with existing manufacturing technology while leveraging Pt's benefits at the critical AOS interface. | | | | deposition along with Pt. However, while Pt exhibits hydrogen activation, its hydrogen storage and transport capabilities are low, leaving it unclear how hydrogen can be transported to the Pt electrode. |
| C22C | A noble metal-based alloy, specifically a platinum alloy (C22C 5/04), can be used to form a conductive interlayer between AOS and electrodes. Platinum alloys are renowned for their superior conductive properties and resistance to oxidation, which is critical in reducing contact resistance. The layer can be deposited using atomic layer deposition, a technique that ensures high precision and conformity, necessary for buried contacts in complex storage architectures. This approach not only addresses the resistance issue but also improves the overall reliability and performance of the storage device. | 3 | 0 | 7 | Using atomic layer deposition (ALD) as a process to apply Pt film to complex structures is a viable possibility. However, while Pt exhibits hydrogen activation properties, its hydrogen storage and transport capabilities are low, making it unclear how hydrogen can be effectively transported to the Pt electrode. |
| G04G | Implementing a hybrid contact approach that combines metal and conductive polymers could address the high contact resistance in AOS-based TFTs. Conductive polymers can conform to the surface irregularities of the AOS, ensuring a more uniform contact and reducing resistance. The metal component provides the necessary conductivity, while the polymer layer enables flexibility and adaptability to the interface. This hybrid structure can be fabricated using a combination of spin coating and metal deposition techniques, allowing precise control over the layer's thickness and composition. This solution is inspired by G04G 9/04, which involves controlling light sources like LEDs, as it utilizes a hybrid approach to optimize performance and efficiency in electronic devices. | 2 | 0 | 2 | Conductive polymers are not applicable to ultra-fine structures. |
| Ni | he hydrogen release property of nickel when reacting with acids can be used to introduce hydrogen into the AOS interface. A controlled acidic environment can be created where nickel, in contact with the AOS, releases hydrogen, leading to increased charge carrier density. This method is efficient for buried contacts as it doesn't require direct exposure to the surface, making it suitable for complex device architectures. | 2 | 0 | 2 | Such a process not applicable to ultra-fine structures. |
| H | Utilize hydrogen-passivated contact surfaces by treating the AOS with hydrogen radicals during deposition. This passivation can prevent charge trapping and improve the | 2 | 0 | 3 | The entire channel layer becomes low-resistance, rendering it incapable of switching. |
| | interface quality between the AOS and electrodes, thereby minimizing contact resistance. The method utilizes hydrogen's reactivity to enhance interface characteristics without adding significant complexity to the fabrication process. | | | | |
| F16F | 1. Adapt the concept of vibration damping from F16F 9/00 by integrating a fluid-based contact layer. This layer will act as a dynamic interface between the AOS and electrodes, compensating for contact resistance variations. The fluid medium, possibly a highly conductive liquid metal, can adjust its position dynamically, ensuring consistent contact and reducing resistance. The adaptability of this fluid layer can help accommodate the vertical stacking of TFTs, maintaining low contact resistance even in complex architectures. | 1 | 0 | 1 | The description is unclear or lacks sufficient information to make a judgment. |
| Co | Leverage cobalt's magnetic properties for contact resistance reduction. By applying an external magnetic field during the deposition process of cobalt-containing layers, the alignment of cobalt atoms can be controlled, possibly enhancing electron flow at the contact interface. This magnetic field-assisted deposition can create more efficient pathways for charge carriers, thereby lowering contact resistance. This innovative approach can be integrated with existing manufacturing processes to enhance TFT performance. | 1 | 0 | 3 | The entire channel layer becomes low resistance, resulting in a loss of switching functionality. |
| Se | Utilizing selenium's photoconductivity, an innovative approach would involve designing AOS TFTs that can locally increase conductivity using targeted light exposure. By embedding selenium in strategic locations within the AOS layer, manufacturers can design TFTs that respond to specific wavelengths of light, enhancing the conductivity precisely where needed. This could be particularly advantageous for buried contacts, as light can be used to selectively excite selenium to reduce resistance without mechanical or thermal interventions. | 1 | 0 | 3 | The reason for the reduction in contact resistance is unclear. |

Table S3. Prompts for generating explanatory descriptions.

| List | SYSTEM_PROMPT: | USER_PROMPT: | Remark |
|---|---|---|---|
| Subclass of IPC | You are an expert in International Patent Classification (IPC). | Please explain what kind of technology is classified under the subclass and title: {IPC_subclass}. When writing your explanation, please also refer to the following main groups or subgroups that exist below the subclass: {IPC_group}. Additionally, consider the hierarchical structure of these groups and subgroups. Please provide a summary in approximately 300 words. | IPC_subclass: The symbol, title, and reference of a single IPC subclass. IPC_group: The symbols, titles, and references of all groups belonging to the specified IPC subclass. |
| Chemical elements | You are an expert in {use_Knowledge}. | Provide a detailed explanation of the scientific properties of {use_Knowledge}, including its key characteristics. Additionally, outline the technologies associated with {use_Knowledge}. Please summarize the information in approximately 300 words. | use_Knowledge: symbol of chemical element |

Table S4. Prompts for generating expert and solutions.

| Method | SYSTEM_PROMPT: | USER_PROMPT: | Remark |
|---|---|---|---|
| Standard | - | Your task is the following: "{Request}". Please create {n} solutions to solve your task. Each solution must be written in about 150 words. | The description of the challenging problem is assigned to "Request." In this study, n=10. |
| SELLM (IPC) | You are an expert in a technology classified under {use_Knowledge} of the International Patent Classification (IPC). Please provide detailed responses to the user's request from the perspective of your expertise. | Your task is the following: "{Request}" Use the following IPC code to solve your task: "{use_Knowledge}". The explanation of the IPC code is as follows: "{use_Knowledge_EXP}"<br><br>Please generate {n} solutions to solve your task. Each solution must be written in about 150 words.<br><br>When you write them down, please include the IPC code that you have used. | The description of the challenging problem is assigned to "Request." use_Knowledge: The area of expertise (subclass) of the LLM expert. use_Knowledge_EXP: An explanation of the technology within the expert's area. In this study, n=10. |
| SELLM (element) | You are an expert in {use_Knowledge}. Please provide detailed responses to the user's request from the perspective of your expertise. | Your task is the following: "{Request}" Use the following element and the technology associated with it to solve your task: "{use_Knowledge}". The explanation of the element is as follows: "{use_Knowledge_EXP}". Please create {n} solutions to solve your task. Each solution must be written in about 150 words. | The description of the challenging problem is assigned to "Request." use_Knowledge: The area of expertise (element) of the LLM expert. use_Knowledge_EXP: An explanation of the technology within the expert's area. In this study, n=10. |

Table S5. Prompt for evaluating the similarity between generated solutions and reference solutions.

| SYSTEM_PROMPT: | USER_PROMPT: | Remark |
| --- | --- | --- |
| You are a strict evaluator. Please conduct a rigorous evaluation by following the described rules while referring to the scoring examples, which consist of solutions, corresponding scores, and comments. | You will now evaluate how closely the generated solutions align with the correct solution. First, the correct solution is defined as "{Desired_Ideas}" When you evaluate it, please consider the following scoring points: "{Points_of_scoring}" The example list of generated solutions, their scores, and comments, are presented below: "{Example_of_scoring}" Using these evaluations as a reference, rigorously assess the {n} generated solutions: "{Generated_Ideas}" Scoring should be conducted on a scale of 1 to 10 based on how closely the generated solutions match the correct solution, and you must provide an explanation for your judgment. If the generated solution lacks elements present in the correct solution or includes irrelevant elements not found in the correct solution, assign a low score (e.g., 1-3). If the solution includes most of the elements of the correct solution and closely resembles it, assign a high score (e.g., 8-10). Please ensure to evaluate each solution individually. | Desired_Ideas: The solutions described in the original literature (reference solution) Points_of_scoring: Key points to consider when assessing similarity. Example_of_scoring: Scoring examples assessed by human experts. Generated_Ideas: The solutions to be evaluated (in this case, 10 solutions each). |

Table S6. Scoring for SBE similarity assessment related to the light extraction problem.

| Solution | HBE | Comment by human expert |
|---|---|---|
| Explore the possibility of embedding high refractive index microspheres, such as those made from titanium oxide, in a thin polymer layer on the glass substrate. This method can provide effective light scattering and maintain surface smoothness by carefully controlling the particle size and distribution. | 3 | For conventional resins with a refractive index of approximately 1.5, light does not easily reach the scattering layer due to total internal reflection at the interface between the transparent electrode and the scattering layer, resulting in low light extraction efficiency. |
| Employ atomic layer deposition (ALD) to create a precisely controlled, thin, and smooth layer of high refractive index material (e.g., hafnium oxide) on the glass substrate. ALD provides excellent uniformity and surface smoothness. | 2 | In this structure, light does not be scattered, leading to low extraction efficiency. |
| Fabricate a composite glass layer using a mixture of high refractive index glass frits and polymer binders which are then densified together to form a smooth, scattering surface. | 6 | It is unclear what is meant by 'densification.' If the process only involves mixing and then applying pressure, the polymer binder with a lower refractive index would be exposed on the surface, which would limit improvements in light extraction. |
| Use a specialized screen printing technique to deposit a mixture of high-refractive-index nanoparticles and a binder onto the glass substrate. The nanoparticles can be carefully chosen to have a refractive index of at least 1.9, and the resulting layer can be cured to form a smooth, even coating that enhances light scattering. | 3 | For conventional resins with a refractive index of approximately 1.5, light does not easily reach the scattering layer due to total internal reflection at the interface between the transparent electrode and the scattering layer, resulting in low light extraction efficiency. |
| Apply a spin-coating technique using a high refractive index polymer mixed with nanoparticles to form a composite scattering layer. The nanoparticles can be selected to achieve the desired refractive index, while the spin-coating ensures a smooth surface. | 4 | The material is not glass, but the concept is similar to KIWI, as it merely replaces high-refractive-index glass with a high-refractive-index polymer. However, it lacks durability. |
| Use layer-by-layer assembly of high refractive index materials such as hafnium oxide (HfO2) combined with organic layers to achieve a smooth, high refractive index scattering layer on the glass substrate. | 2 | In this structure, light does not be scattered, leading to low extraction efficiency. |
| Manufacture a glass substrate with embedded high refractive index particles, such as zirconium dioxide (ZrO2), during the glass forming stage. This approach can increase the overall refractive index of the substrate itself, enhancing light extraction. | 6 | While it is possible to form a glass scattering layer, maintaining surface smoothness means that the matrix of the scattering layer will be formed by the glass substrate. As the refractive index of a typical glass substrate is around 1.5, improvements in light extraction efficiency are limited. |
| Use a phase-separated glass approach where during the cooling process, phase separation creates microdomains of different refractive indices enhancing light scattering and achieving a smooth surface after annealing. | 3 | Using phase-separated glass is an interesting idea, but it is unclear how to form a matrix of high refractive index glass. |
| Apply a high-index glass frit paste onto the substrate through screen printing and then sinter it to form a continuous layer. The screen printing process is categorized under B05D, and the glass sintering is part of C03B. | 9 | This refers to the process of forming a high refractive index glass material necessary for light extraction and forming a high refractive index glass scattering layer, using high refractive index glass frit as a raw material in pattern formation through screen printing and scattering layer formation through firing. Although no scattering particles are included, the presence of bubbles remaining in the fired glass layer can act as substitutes for scattering particles, thus significant improvements in light extraction efficiency are expected. |
| Use sol-gel processing to create a thin, uniform ceramic coating with high refractive index on the glass substrate. The sol-gel process allows precise control over the material composition and surface smoothness, which is crucial for preventing electrode short circuits. | 3 | Although the sol-gel ceramic layer mentioned is glass, the overall reproduction quality is low. |
| Formulate a glass frit composition with added high-refractive-index materials and apply it as a thin film on the substrate. The glass frit can be sintered to form a smooth and uniform scattering layer. | 9 | Although it uses frit rather than frit paste, it can be regarded as a creative solution for an early idea. |
| Combine glass frits with high-refractive-index additives and apply them as a paste to the substrate. The paste can be sintered to form a smooth, continuous layer. This method requires careful formulation to achieve the desired optical properties. | 7 | The refractive index of glass frit is not limited to a high refractive index, so in order to increase light out-coupling efficiency, the high refractive index additive must contain more than the glass frit. For example, when high-refractive-index ceramics are used as additives, the amount of glass may be small and difficult to sinter, or even if sintered, ceramic particles may be exposed on the surface and a smooth surface cannot be obtained. |
| Develop a ceramic-based layer using high refractive index materials like titanium dioxide (TiO2) or zirconium dioxide (ZrO2) mixed with glass frits. These materials can be sintered to form a smooth, dense layer on the glass substrate. The ceramic approach ensures the layer is both smooth and durable while having a high refractive index. | 7 | The refractive index of glass frit is not limited to a high refractive index, so in order to increase light out-coupling efficiency, the high refractive index additive must contain more than the glass frit. For example, when high-refractive-index ceramics are used as additives, the amount of glass may be small and difficult to sinter, or even if sintered, ceramic particles may be exposed on the surface and a smooth surface cannot be obtained. |
| Utilize a co-sintering process to combine high-refractive-index glass frits with standard glass. The frits are designed to have a refractive index greater than 1.9, and the co-sintering ensures a smooth, uniform surface. | 10 | By carefully selecting the particle size and softening temperature of glass powders, it is possible to co-sinter high refractive index glass powder with standard refractive index glass powder (~1.5) to obtain a flat-surfaced glass scattering layer. |
| Employ the screen printing technique to apply a layer of glass paste mixed with high-refractive index materials onto | 7 | The refractive index of glass frit is not limited to a high refractive index, so in |
| the substrate. The paste can be formulated to have a fine particle size and a high TiO2 content to achieve the desired refractive index. After printing, the layer is sintered to form a smooth glassy surface that adheres well to the substrate. | | order to increase light out-coupling efficiency, the high refractive index additive must contain more than the glass frit. For example, when high-refractive-index ceramics are used as additives, the amount of glass may be small and difficult to sinter, or even if sintered, ceramic particles may be exposed on the surface and a smooth surface cannot be obtained. |
| Incorporate a high-refractive index glass frit, such as lead glass, into a slurry that is screen printed onto the substrate. The frit is then sintered to form a continuous smooth layer with enhanced refractive index properties suitable for light scattering. | 9 | This refers to the process of forming a high refractive index glass material necessary for light extraction and forming a high refractive index glass scattering layer, using high refractive index glass frit as a raw material in pattern formation through screen printing and scattering layer formation through firing. Although no scattering particles are included, the presence of bubbles remaining in the fired glass layer can act as substitutes for scattering particles, thus significant improvements in light extraction efficiency are expected. |
| Use a tape-casting method to create a uniform layer of glass containing light-scattering particles. The glass slurry can be cast onto the substrate and fired to form a dense and smooth surface with high refractive index. | 6 | It says it will have a high refractive index, but why that would be the case is not readily apparent. |
| Apply a layer of high refractive index glass frit mixed with titania onto the substrate, followed by a sintering process to form a smooth, dense layer that provides scattering properties. | 10 | It satisfies all the requirements necessary to address the issue and successfully reproduces the creative solution. |
| Adopt a hot pressing method to form a composite layer of glass frits and high-refractive-index particles, ensuring uniformity in particle dispersion and achieving a smooth surface through controlled pressing. | 6 | The hot pressing method can be used, but it does not say that the glass has a high refractive index. |
| Adopt a screen printing technique to apply a paste composed of high-refractive-index materials like SiO2 doped with TiO2. This allows for the creation of a uniform layer, as screen printing can be used to accurately control the thickness and surface quality of the layer. | 5 | The content is incomplete and cannot be considered a proposed solution. Ti does not incorporate significantly into SiO$_2$, and there is no description of the process after printing. |

Table S7. Scoring for SBE similarity assessment for the problem of contact resistance in IGZO-TFT.

| Solution | HBE | Comment by human expert |
|---|---|---|
| Another solution is to employ a plasma-enhanced chemical vapor deposition (PECVD) process to deposit a nanolayer of catalytic nanoparticles on the AOS contact surfaces. These nanoparticles, possibly composed of platinum or palladium, are known for their catalytic efficiency in hydrogen dissociation. By integrating these into the contact design, they can facilitate the activation and diffusion of hydrogen species into the AOS, reducing contact resistance. This method aligns with B01J 8/00, which deals with fluid-solid interactions, as it involves the interaction of gaseous hydrogen with solid catalytic particles to enhance conductivity at the interface. This deposition technique can be fine-tuned to form a uniform layer even on buried contacts within vertically stacked architectures, providing a scalable solution to the contact resistance problem. | 2 | Point A is not included. Regarding Point B, it is included, but using PECVD to deposit platinum or palladium nanoparticles does not seem to be feasible. |
| An innovative approach could involve creating a graded interface using atomic layer deposition (ALD) to apply a thin layer of an amorphous conductive oxide, such as zinc indium oxide (ZIO), which is doped with a catalytic metal like palladium. The ALD process, related to B01J 21/00 for catalysis, enables atomic-level precision in layer deposition, ensuring excellent conformity and coverage even on buried interfaces. The catalytic doping facilitates hydrogen activation at the interface, thus enhancing charge carrier mobility and reducing contact resistance. This graded interface can be tailored to optimize performance across different contact depths and complexities within the device. | 3 | Point A is not included. Pd doping over the entire AOS surface can degrade the transistor characteristics. |
| Developing an electrode with a self-assembled monolayer (SAM) of catalytic molecules, such as those containing palladium or rhodium, can provide a direct solution to the contact resistance issue. These SAMs, applied using B01J 41/00 ion-exchange techniques, create a uniform and highly catalytic surface on the electrode, promoting efficient hydrogen dissociation and charge transfer. This method ensures that even deep, buried contacts are effectively treated, as the monolayer can conform to the underlying surface, maintaining low resistance across various device configurations. The technique's simplicity and adaptability make it suitable for integration into existing manufacturing processes for advanced storage devices. | 2 | Point A is not included. Pd is included in the output. However, it can not function as a transistor if the SAM film does not have its own conductivity. |
| To address the high contact resistance in stacked AOS TFTs, leverage B01J's catalysis processes to develop a catalytic interlayer that reduces contact resistance. By designing a nanoscale interlayer composed of metal oxides such as indium tin oxide (ITO) doped with catalytic metals like platinum or palladium, we can enhance electron transfer between AOSs and electrodes. The catalytic activity of these metals can be tuned to facilitate the dissociation of hydrogen molecules into atomic hydrogen at the contact interface, which, in turn, improves charge carrier injection by creating more conductive pathways. This approach not only reduces contact resistance effectively but can | 3 | Point A is not included. Atomic Pd doped into ITO may not have the catalytic effect. |
| also be adapted for buried contacts within complex architectures due to the nanoscale thickness and high activity of the catalytic interlayer, ensuring minimal interference with device operation. | | |
| One approach to address the contact resistance issue in AOS TFTs could involve the development of a novel catalytic interface layer using a mixed metal oxide catalyst tailored to enhance hydrogen dissociation. By leveraging B01J's focus on catalysis, specifically metal oxide catalysts, a thin film of a catalyst such as platinum-doped tin oxide could be deposited at the AOS-electrode interface. This layer would facilitate the dissociation of hydrogen molecules into atomic hydrogen, which could effectively reduce contact resistance by increasing charge carrier density. The thin catalytic layer could be engineered to be conductive itself, providing dual functionality as a contact interlayer, enhancing both electrical contact and hydrogen activation. | 3 | Pd-doped tin oxide may not have the catalytic effect. |
| The idea is to employ a composite electrode material that incorporates a catalytic layer to lower contact resistance at the AOS-electrode interface. By using a catalytic oxide such as titanium oxide doped with elements like platinum or palladium, we can enhance the generation of charge carriers at the interface. The catalytic layer can be engineered to create oxygen vacancies in situ, thereby facilitating better charge transfer between AOS and the electrode. This approach leverages the catalytic properties under B01J 23/00, which focuses on catalysts comprising metals or their compounds, to improve electrical contact performance in deeply buried contacts. | 2 | Palladium-doped titanium is not conductive, so it cannot be used as an electrode material. |
| An effective solution for reducing contact resistance in AOS TFTs is the use of a metallic nanowire interlayer. Metallic nanowires, such as silver or copper, offer high conductivity and can be integrated into the device architecture to provide a conductive bridge between the AOS and the electrodes. The process involves depositing a network of metallic nanowires onto the AOS surface using techniques such as spin-coating or inkjet printing, followed by thermal annealing to improve contact and conductivity. The electrodes can be fabricated using metals that form reliable connections with the nanowires, such as gold or palladium. This nanowire interlayer enhances carrier injection and reduces contact resistance, while also maintaining the transparency and flexibility of the device. This method is particularly advantageous for buried contacts in vertically stacked architectures, as it provides a scalable and efficient solution to the contact resistance issue, compatible with existing manufacturing processes. | 1 | "Spin coating or inkjet cannot be applied to the embedded structures. |
| One potential solution is to use a novel composite material that combines amorphous oxide semiconductors (AOS) with a thin layer of graphene. Graphene's outstanding electrical conductivity and flexibility make it a suitable candidate to act as a bridge between the AOS and the electrodes. The proposed approach involves depositing a thin graphene layer onto the AOS surface using chemical vapor deposition (CVD). This graphene layer can help to significantly lower the contact resistance by providing a | 1 | IGZO cannot be reduced using nanowires." |
| highly conductive pathway for charge carriers. Furthermore, the graphene layer can protect the AOS surface from degradation during device fabrication and operation. The electrodes can be made of a metal with good adhesion to graphene, such as palladium or nickel, which would ensure strong contact and minimize resistance. The process could be further optimized by functionalizing the graphene surface to enhance its interaction with the AOS, possibly by introducing dopants or defects. This method aims to address the contact resistance issue effectively while maintaining compatibility with current manufacturing processes for high-density storage devices. | | |
| To address the challenge of high contact resistance in AOS TFTs, a viable solution is the incorporation of a highly conductive metal oxide heterostructure, such as a bilayer of titanium oxide (TiO2) and indium oxide (In2O3). These materials can be engineered to exhibit complementary electronic properties, optimizing the conductive path. The process involves sequential deposition of TiO2 and In2O3 layers onto the AOS surface using sputtering or atomic layer deposition (ALD). The electrodes can be fabricated using metals like aluminum or nickel, which form good contacts with these metal oxides. The heterostructure serves to enhance charge carrier injection and reduce contact resistance by providing a graded energy band alignment at the interface. Furthermore, this configuration can be tailored to improve stability and environmental resistance. This approach offers a scalable and effective solution for reducing contact resistance in vertically stacked architectures, aligning with the requirements of next-generation high-density storage devices. | 3 | Graphene must be formed only in the region where the source-drain electrode and IGZO contact each other, or it will not work as a transistor. And even if it could, the IGZO cannot be reduced. |
| To address the high contact resistance in AOS TFTs, a layered electrode structure using a copper (Cu) core with a graphene interlayer can be implemented. Graphene, known for its exceptional electrical conductivity and ultra-thin profile, can be used to create an effective conductive interlayer between the copper electrode and the AOS. The process involves depositing a thin layer of graphene on the AOS surface using chemical vapor deposition (CVD), followed by the sputtering of copper to form the electrode. This approach leverages the low resistivity of copper and the high electron mobility of graphene to facilitate efficient charge transfer. Additionally, graphene's flexibility and chemical stability make it ideal for integration into vertically stacked architectures, minimizing energy requirements by reducing the need for complex processing steps. This method not only lowers contact resistance but also provides a scalable solution for buried contacts in complex memory device structures. By optimizing the graphene layer thickness and ensuring a defect-free interface, this method can significantly enhance the performance and efficiency of AOS-based TFT storage devices. | 2 | The idea of stacking multiple materials to get a gradient in energy is interesting. However, IGZO cannot be reduced and low contact resistance cannot be obtained. |
| Utilize palladium as an interfacial layer between the AOS and the electrode. Due to its excellent hydrogen absorption properties, Pd can effectively dissociate hydrogen molecules into atomic hydrogen, which can then penetrate the AOS to reduce contact | 8 | It cannot be applied to the embedded TFTs. Also, IGZO cannot be reduced by graphene, so low contact resistance cannot be obtained. |
| resistance. The high hydrogen solubility of Pd facilitates the generation of charge carriers that mitigate the contact resistance without requiring additional energy-intensive processes. Moreover, the chemical stability and resistance to oxidation of palladium ensure that the interfacial layer remains effective over time, even in the vertically stacked configurations of advanced storage devices. | | |
| Design a heterojunction using palladium and an amorphous oxide semiconductor, where Pd acts as a catalytic layer for hydrogen dissociation. This heterostructure can be engineered to maximize hydrogen solubility and charge carrier generation at the interface, improving the electrical characteristics of the TFTs. The heterojunction approach can be tailored to achieve a balance between electronic performance and structural integrity, essential for high-density DRAM applications. | 8 | It is good that Pd is used as the interfacial layer. On the other hand, there is no description of electrode materials other than the interface layer. |
| Use a Pd-based alloy as the electrode material to inherently possess excellent hydrogen absorption capabilities. An alloy such as Pd-Cu or Pd-Ag can be tailored to maintain high conductivity while enhancing hydrogen diffusion and dissociation at the interface with AOS. The alloy's properties can be optimized to ensure minimal contact resistance without necessitating additional surface treatments or energy-intensive processes, making it suitable for buried contacts in high-density memory applications | 9 | The design concept of heterojunction is good. However, there is no description of electrodes. |
| Employ a palladium-titanium bilayer electrode structure. The Pd layer provides hydrogen absorption and diffusion, while the Ti layer ensures strong adhesion and low electrical resistance. This bilayer structure is deposited sequentially using electron-beam evaporation, allowing precise control over layer thickness and uniformity. The Pd layer serves as a continuous hydrogen source, reducing contact resistance by promoting hydrogen diffusion into the AOS. This approach is particularly effective for vertically stacked devices, as the bilayer structure accommodates the mechanical and thermal stresses inherent in complex architectures. | 8 | This approach attempts to combine the ability to reduce IGZO by hydrogen using Pd as a catalyst and its conductivity as an electrode. It could be a solution. |
| To effectively reduce contact resistance in AOS TFTs, one can utilize a zinc oxide (ZnO) nanorod interfacial layer. ZnO nanorods have been shown to provide excellent electrical contact properties due to their high aspect ratio and conductive nature. The fabrication process involves the hydrothermal growth of ZnO nanorods directly on the AOS surface, followed by the deposition of a metal electrode such as aluminum (Al) via thermal evaporation. The ZnO nanorods serve as a bridging layer that enhances charge transfer by providing a high-density conductive network at the interface, thus minimizing contact resistance. Furthermore, the surface of the ZnO nanorods can be functionalized with hydrogen plasma treatment to create additional charge carriers, further improving conductivity. This approach is particularly suitable for buried contacts in vertically integrated architectures, as the nanorods can effectively penetrate and conform to complex geometries. The use of ZnO nanorods is energy-efficient and can be easily integrated into existing | 1 | This approach attempts to combine the ability to reduce IGZO by hydrogen using Pd as a catalyst and the conductivity of Ti as an electrode, and could be a solution. However, the use of Ti is counterproductive because the specific resistance of Pd is 10.8 μΩ and that of Ti is 55 μΩ, which is rather higher than that of Pd. |
| semiconductor processes, providing a scalable solution for high-density storage applications. | | |
| A novel solution involves using a cobalt oxide (CoO) nanosheet as an interfacial layer between the AOS and the metal electrode. Cobalt oxide has been shown to possess excellent electron transport properties and can form a low-resistance ohmic contact with AOS materials. The process includes the hydrothermal synthesis of CoO nanosheets, which are then transferred onto the AOS layer. Following this, a metal such as gold (Au) is deposited on top of the CoO layer using electron-beam evaporation. The CoO nanosheet acts as an effective buffer layer that reduces contact resistance by facilitating charge transfer across the interface and minimizing the formation of interface traps. Moreover, CoO can be engineered to have oxygen vacancies, enhancing its conductivity without the need for additional surface treatments. This method is particularly suited for buried contacts in complex 3D architectures, as the nanosheets can self-assemble into the desired configuration. Additionally, the use of CoO is compatible with existing semiconductor processing techniques, making it a practical choice for integration into current manufacturing workflows. | 1 | Not applicable because IGZO is damaged during hydrothermal synthesis. |
| An innovative approach to mitigate contact resistance in AOS TFTs is the use of a titanium nitride (TiN) interlayer. TiN is known for its metallic conductivity and excellent thermal stability, making it suitable for high-density DRAM applications. The process involves depositing a thin TiN layer onto the AOS using reactive sputtering, followed by the deposition of a copper electrode. TiN serves as a conductive bridge that facilitates efficient charge transfer between the AOS and the metal electrode. Additionally, TiN can be treated with nitrogen plasma to create a nitrogen-rich surface, reducing interface states and improving contact adhesion. This method is advantageous for buried contacts, as TiN's robustness ensures sustained performance even under high thermal and electrical stress. The use of TiN not only reduces contact resistance but also provides a diffusion barrier, preventing metal diffusion into the AOS layer. This approach simplifies the manufacturing process by eliminating the need for complex surface treatments and is compatible with large-scale production techniques, making it ideal for next-generation storage device architectures. | 4 | Even if the structure is created, it will not have low contact resistance because IGZO cannot be reduced." |
| A promising solution for reducing contact resistance in AOS TFTs is the use of nanoscale metal catalysts embedded within the AOS matrix at the contact interface. These metal nanoparticles, such as palladium or platinum, can be synthesized via colloidal chemistry techniques (B01J 13/00), which allow for precise control over particle size and distribution. The embedded catalysts promote hydrogen dissociation at the nanoscale, enhancing charge carrier generation directly at the contact points. This nanoscale approach ensures that even the most challenging buried contacts in vertically stacked architectures benefit from reduced resistance. By integrating these metal nanoparticles into the AOS material, the | 2 | Not applicable because IGZO is damaged during hydrothermal synthesis. |
| method capitalizes on the catalytic properties of the metals while maintaining the integrity and performance of the semiconductor. | | |
| **Hydrogen Plasma Treatment with Nanostructured Catalysts:** Introduce nanostructured catalysts, such as Pd or Pt nanoparticles, at the AOS-electrode interface to enhance the efficiency of hydrogen plasma treatments. These catalysts can promote the dissociation of hydrogen molecules and facilitate charge carrier generation, thereby reducing contact resistance even at buried interfaces where direct access is limited. | 1 | Introducing nanoparticle catalysts at the buried AOS-electrode interface does not allow exposure to hydrogen plasma because of the embedded structure. |
| Design a composite electrode consisting of palladium and a conductive oxide, such as indium tin oxide (ITO), to improve contact resistance in TFTs. The palladium component can absorb hydrogen and release atomic hydrogen at the interface, thereby increasing charge carrier density and reducing resistance. The conductive oxide can provide a stable, high-conductivity matrix that ensures good electrical contact and mechanical stability. This combination can be particularly effective in vertically stacked architectures where the palladium can address contact resistance issues, while the conductive oxide ensures that the electrode maintains its integrity and performance under complex operational conditions. | 8 | This approach attempts to combine the ability to reduce IGZO by hydrogen using Pd as a catalyst and the conductivity of ITO as an electrode, and could be a solution. However, the use of ITO is counterproductive because the specific resistance of Pd is 10.8 $\mu\Omega$ and that of ITO is 200 $\mu\Omega$, which is rather higher than that of Pd. |